\documentclass[sigconf]{acmart}
\pdfoutput=1
\usepackage{hyperref}
\usepackage{graphicx}
\usepackage{subcaption}
\AtBeginDocument{%
  }

\setcopyright{acmcopyright}
\copyrightyear{2022}
\acmYear{2022}
\acmDOI{XXXXXXX.XXXXXXX}

\acmPrice{}
\acmISBN{}
\acmDOI{}

\acmConference[KDD-UC'22]{ KDD Undergraduate Consortium}{August 14--18, 2022}{Washington, D.C.}

\begin{document}

\title{Challenges and opportunities in applying Neural Temporal Point Processes to large scale industry data}

\author{Dominykas Šeputis}
\email{dominykas.seputis@mif.stud.vu.lt}
\affiliation{%
  \institution{Vilnius University}
  \streetaddress{Naugarduko g. 24}
  \city{Vilnius}
  \country{Lithuania}
  \postcode{03225}
}

\author{Jevgenij Gamper}
\email{jevgenij.gamper@vinted.com}
\affiliation{%
  \institution{Vinted}
  \streetaddress{Švitrigailos g. 13-1-4 floor}
  \city{Vilnius}
  \country{Lithuania}
  \postcode{03228}
}

\author{Remigijus Paulavičius}
\email{remigijus.paulavicius@mif.vu.lt}
\affiliation{%
  \institution{Vilnius University}
  \streetaddress{Naugarduko g. 24}
  \city{Vilnius}
  \country{Lithuania}
  \postcode{03225}
}

\renewcommand{\shortauthors}{Šeputis et al.}

\begin{abstract}
In this work, we identify open research opportunities in applying Neural Temporal Point Process (NTPP) models to industry scale customer behavior data by carefully reproducing NTPP models published up to date on known literature benchmarks as well as applying NTPP models to a novel, real world consumer behavior dataset that is twice as large as the largest publicly available NTPP benchmark. We identify the following challenges. First, NTPP models, albeit their generative nature, remain vulnerable to dataset imbalances and cannot forecast rare events. Second, NTPP models based on stochastic differential equations, despite their theoretical appeal and leading performance on literature benchmarks, do not scale easily to large industry-scale data. The former is in light of previously made observations on deep generative models. Additionally, to combat a cold-start problem, we explore a novel addition to NTPP models - a parametrization based on static user features. 
\end{abstract}


\maketitle

\section{Introduction}
Considering online platforms, most use cases of generative modeling applications are motivated by the need to capture user behavior. When in need to answer questions such as "How long do we need to wait for the next user's visit?" or "What is the optimal time of recommending winter coats?" prediction type tasks where the goal is to predict times and marks of the subsequent events are formed. With the rapid growth of online platforms, it is crucial to estimate how its users will behave in the near future.

Commonly referred to as Buy-till-You-Die models, Pareto/NBD \cite{10.5555/2780513.2780514}, BG/NBD \cite{10.5555/2882600.2882608}, and other generative models alike \cite{jasek_comparative_2019} might be too simplistic and lack flexibility in modeling heterogeneity of customer behavior. Nonetheless, the generative approach towards customer behavior modeling described by \cite{FADER200961} might be useful, albeit in a different form.

For example, in Pareto/NBD model, the number of purchases that an $i’\text{th}$ customer makes in their lifetime is described by a Poisson distribution parameterized by $\lambda_i$. Given the inferred parameters for Poisson and Exponential distributions representing a customer's purchase number and lifetime, one can estimate the probability of a customer being "alive." Buy-till-You-Die models assume a fixed parameter $\lambda$ that does not depend on the time, and the models are static. A customer who had a positive experience with a service is more likely to come back and keep returning if the experience continues to be positive and vice versa - these models do not capture such behavior. 

An alternative would be to use a self-exciting point process like Hawkes, which implies that $\lambda$ parameter is a function of time, positively influenced by past events. Lately, a new class of temporal point processes (TPPs) has emerged - Neural Temporal Point Process (NTPP) \cite{NIPS2017_6463c884, 1907.07561, 1806.07366}. The NTPPs are based on the self-exciting nature of the Hawkes process but are more versatile and can capture more diverse user behavior. In this work, we identify open research opportunities in applying neural temporal point process models to industry scale customer behavior data. 

We explore and compare different approaches to NTPP: Neural Ordinary Differential Equations \cite{1806.07366} and Attention \cite{1706.03762} based temporal point processes \cite{1907.07561, 2007.13794}. We apply these models to various synthetic and real-life datasets. To further test these models, we introduce online second-hand marketplace users behavior dataset. The dataset consists of 4,886,657 events and is 10 times larger than the commonly used Stack Overflow dataset \cite{10.1145/2939672.2939875, 1905.10403, 2007.13794, 10.5555/3295222.3295420}. We attempt to use selected models to describe users' actions in the marketplace and explore the benefits of parameterizing Self-Attention based NTPP models with static features. Finally, we discuss explored models' scalability and their ability to capture rare events.

This paper is organized as follows. In section 2, we explore different ways temporal point processes are used in the industry. In section 3, we give a formal overview of traditional and neural temporal point processes. In section 4, we describe the data and our experimental approach. Section 5 reports the main results of the document, and the final section concludes.

\section{Related work}
With the rapid growth of online platforms, it is crucial to estimate how its users will behave in the near future. Multivariate point processes like FastPoint \cite{10.1007/978-3-030-46147-8_28} prove that optimal results are achieved by combining neural networks with traditional generative temporal point process models. The approach scales to millions of correlated marks with superior predictive accuracy by fusing deep recurrent neural networks with Hawkes processes. Combining traditional temporal point process models with neural networks enables the industry to solve diverse problems: from predicting when customers will leave an online platform to simulating A/B tests.

\paragraph{Churn prediction}
Customer churn prediction is one of the most crucial problems to solve in subscription and non-subscription-based online platforms. Traditionally churn prediction is handled as a supervised learning problem \cite{1604.05377, 2201.02463, 1909.11114} where the goal is to predict if a user will be leaving the platform after a fixed time period. However, platform usage habits can differ from one user to another, and a decrease in an activity does not always is a sign of churn. \cite{1909.06868} is one of the solutions that tackle churn prediction via generative modeling. The solution interprets user-generated events in time scope of sessions and proposes using user return times and session durations to predict user churn.

\paragraph{Time-Sensitive Recommendations}
When considering item recommendations for a user, we are thinking about the optimal suggestion and what time of the year we should suggest. \cite{NIPS2015_136f9513} introduces a novel convex formulation of the problems by establishing an under-explored connection between self-exciting point processes and low-rank models. The proposed solution expands its applicability by offering time-sensitive recommendations and users' returning time predictions that can determine previously discussed churn or optimize marketing strategies. The item recommendation part is accomplished by calculating intensity $\lambda_{u,i}$ for each item $i$ and user $u$. After, items are sorted by descending order of calculated intensity, and top-k items are returned.

\paragraph{Interactive Systems Simulation}
It is common to see modern systems using numerous different machine learning models interacting with each other. As a result, user experience is often defined by various machine learning systems layered iteratively atop each other. The previously discussed item-recommendation problem is one example of such complex system \cite{10.1145/2843948}. When there is a need to change one or multiple models in the recommendation system, one might ask themselves what effect a particular change might have on the overall system performance. Usually, A/B testing answers these types of questions. Although when different system modules are changing rapidly and systems are getting more complex, it becomes more practical to simulate the effect than test it using traditional methods. \cite{10.1145/3460231.3474259} presents one of many ways to simulate interactive systems. The paper proposes Accordion, a fully trainable simulator for interactive systems based on inhomogeneous Poisson processes. While combining multiple intensity functions, the Accordion enables a comparison between realistic simulation settings and their effect on the total number of visits, positive interactions, impressions, and any other empirical quantity derived from a sampled dataset. 

\begin{figure}[!ht]
  \centering
    \includegraphics[width=\linewidth]{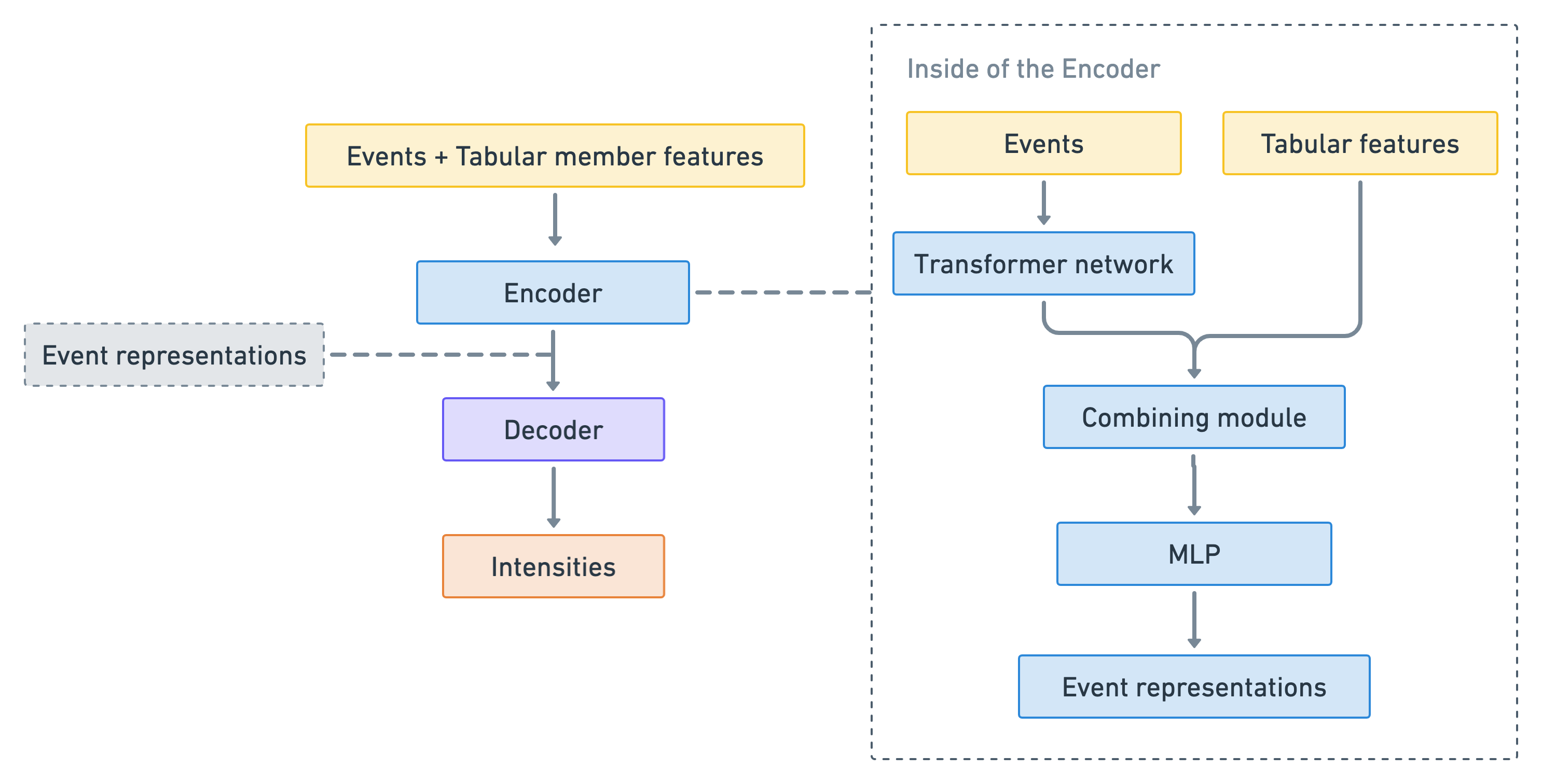}
  \caption{Architecture of parameterized Self Attention encoder.}
  \Description{Architecture of parameterized Self Attention encoder.}
  \label{fig:network_structure}
\end{figure}

\section{Methods}
The base of our explored generative user behavior modeling methods is Poisson process - a continuous-time version of a Bernoulli process where event arrivals are happening at any point of the time \cite{bertsekas2000introduction}. The homogeneous Poisson process describes intensity of event arrival by $\lambda = \lim_{\delta \rightarrow 0} = \frac{P(1, \delta)}{\delta}$, where $\delta$ is a minimal interval between events time. If needed to model events that are influenced by the time dimension, we can use non-homogeneous version of the Poisson process. Simply instead of using a static intensity function $\lambda$ to represent the intensity of the events, we transform it into a function of time $\lambda = \lambda(t)$ \cite{pishro2014introduction}.

To capture more complex behavior, for example item recommendation, where one successful transaction mostly reinforces members to buy more items, self-exciting processes are used. The self-exciting process is a point process with a property that the arrival of the event causes the conditional intensity function to increase. One of the most known self-exciting processes is Hawkes process \cite{10.1093/biomet/58.1.83}. The key feature of the process is a conditional intensity function, influenced by the events in the past $\lambda(t|\mathcal{H}_{t})=\lambda_{0}(t) + \sum_{i:t>T_{j}}\phi (t - T_{j}),$ where $\mathcal{H}_{t}$ is the history of the events that happened before time $t$. From the defined Hawkes function we can see that there is two major parts in it: base intensity function $\lambda_{0}$ and a summation of \textit{kernel} $\phi(t - T_{i})$. The base intensity function $\lambda_{0}$ keeps the independence property from the Poisson process, and past events do not influence it. The influence of the past arrivals in the Hawkes is defined with a kernel function $\phi(\cdot )$, which usually takes form of exponential function $\phi(x) = \alpha e^{-\beta x} \mathcal{,}$ where $\alpha \geq 0 \mathcal{,} \beta > 0$ and $\alpha < \beta$. The kernel function ensures that more recent events would have a greater influence on the current intensity than events in the past. 

\subsection{Neural Hawkes Process}

While Hawkes process brings a significant improvement in capturing the self-exciting type of events, the process is not fully sufficient at representing a more complex type of events. Hawkes process's intensity function dictates additive and decaying influence of the past events. While this behavior can be true for some events, event streams like item recommendations do not benefit from the Hawkes process's additive nature. The more similar items are shown, the less influence they bring and even can discourage people from buying the product. Neural Hawkes Process (NHP) \cite{NIPS2017_6463c884} is better suited to battle more complex TPP problems. The Neural Hawkes Process introduces methods that can define processes where a past event's influence: is not always additive, is not always positive, and does not always decay. The Neural Hawkes process brings improvements in two parts. First, NHP introduces non-linear \textit{transfer function} $f: \mathbb{R} \rightarrow  \mathbb{R}^{+}$, which usually takes form of non-linear \textit{softplus} function $f(x) = \alpha\log(1 + e^{x / \alpha})$. Second, rather than predicting $Hawkes\:\lambda(t)$ as a simple summation, NHP uses recurrent neural network (RNN)  \cite{doi:10.1073/pnas.79.8.2554} to determine the intensity function $\lambda = f(h(x)),$ where $h$ is a hidden state of the RNN. 
This change allows learning a complex dependence of the intensities on past events' number, order, and timing. This architecture lets to improve the base Hawkes process shortcomings: inability to represent events where past event's influence is not always additive is not always positive, and does not always decay; however, the neural Hawkes process has a downside - weaknesses of RNNs are inherited; namely the model's inability to capture long-term dependencies (also called as "forgetting") and is difficult to train \cite{bengio_learning_1994}.

\subsection{Self-Attention Hawkes process}
Despite not being directly applicable to model point process, recently, Transformer architecture was successfully generalized to continuous-time domain \cite{2002.09291, 1907.07561}. Self-Attentive Hawkes Process (SHAP) is an self-attention based approach to a Hawkes process proposed by \cite{1907.07561}. SHAP employs self-attention to measure the influence of historical events to the next event. The SHAP approach to the temporal point processing field also brings a time-shifted positional embedding method where time intervals act as phase shifts of sinusoidal functions. Transformers Hawkes Process (THP) presented by \cite{2002.09291} is another self-attention based approach to the Hawkes process. THP improves recurrent neural network-based point process models that fail to capture short-term and long-term temporal dependencies \cite{hochreiter2001}. The key ingredient of the proposed THP model is the self-attention module. Different from RNNs, the attention mechanism discards recurrent structures. However, the THP model still needs to be aware of the temporal information of inputs, such as timestamps.  To achieve this temporal encoding procedure is proposed that is used to compute the positional embedding.

\subsection{Neural Jump Stochastic Differential Equations}

Sometimes user's interest is built up continuously over time but may also be interrupted by stochastic events. To simultaneously model these continuous and discrete dynamics, we can use the Neural Jump Stochastic Differential Equations (NJSDEs) \cite{1905.10403}. The NJSDE uses latent vector $z(t)$ to encode the system's state, which continuously flows over time until an event at random, introducing an abrupt jump and changing its trajectory. The event conditional intensity function and the influence of the jump are parameterized with neural networks as a function of $z(t)$, while the continuous flow is described by Neural Ordinary Differential Equations \cite{1806.07366}. The advantage of Neural NJSDEs is that they can be used to model a diverse selection of marked point processes, where a discrete value can compliment events (for example, a class of purchased item) or a vector of real-valued features (e.g., spatial locations).

\subsection{Parameterised Self-Attention Hawkes process}
To combat a cold-start problem, we explore a novel addition to an encoder-decoder architecture based NTPP models - a parametrization based on static user features. Specifically, we experiment with self-attention type encoders. We concatenate outputs of the Transformer model $m$ with processed static user features $p$ and pass them thought multilayer perceptron (MLP): $f(X) = MLP(m||p)$. The change differs from the approach proposed by \cite{2007.13794}, where outputs of the Transformer model are directly passed to the MLP. We present the illustrated architecture of our approach in Figure \ref{fig:network_structure}.

\section{Experimental setup}
\subsection{Data}

\begin{figure}[!ht]
  \centering
    \begin{subfigure}{\linewidth}
      \includegraphics[width=\linewidth]{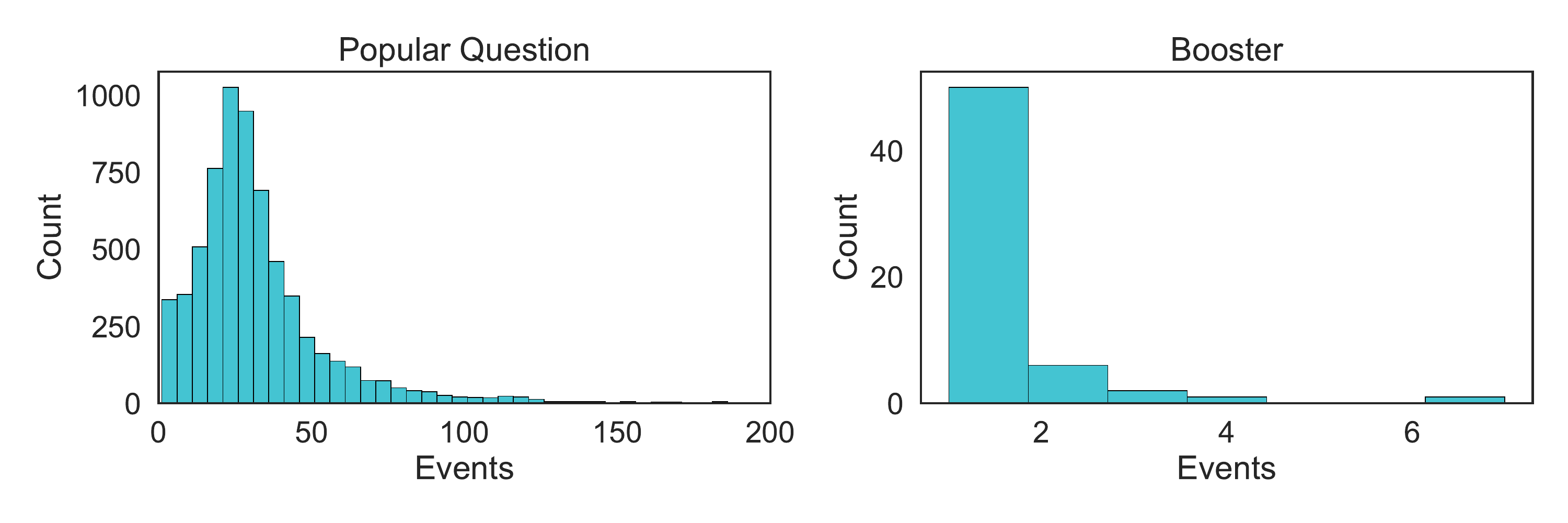}
      \caption{Comparison between common (left) and rare (right) event type distribution within Stack Overflow platform.}
    \end{subfigure}
  \begin{subfigure}{\linewidth}
      \includegraphics[width=\linewidth]{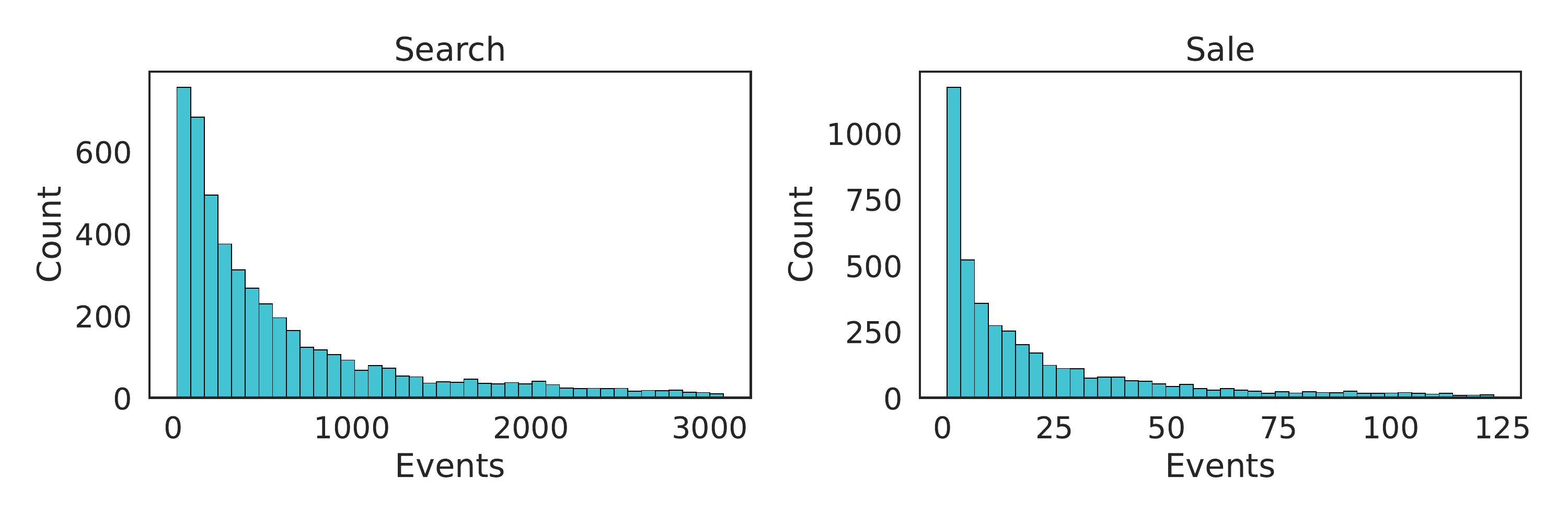}
      \caption{Comparison between common (left) and rare (right) event type distribution within Vinted platform.}
    \end{subfigure}
      \begin{subfigure}{\linewidth}
        \centering      
      \includegraphics[width=0.5\linewidth]{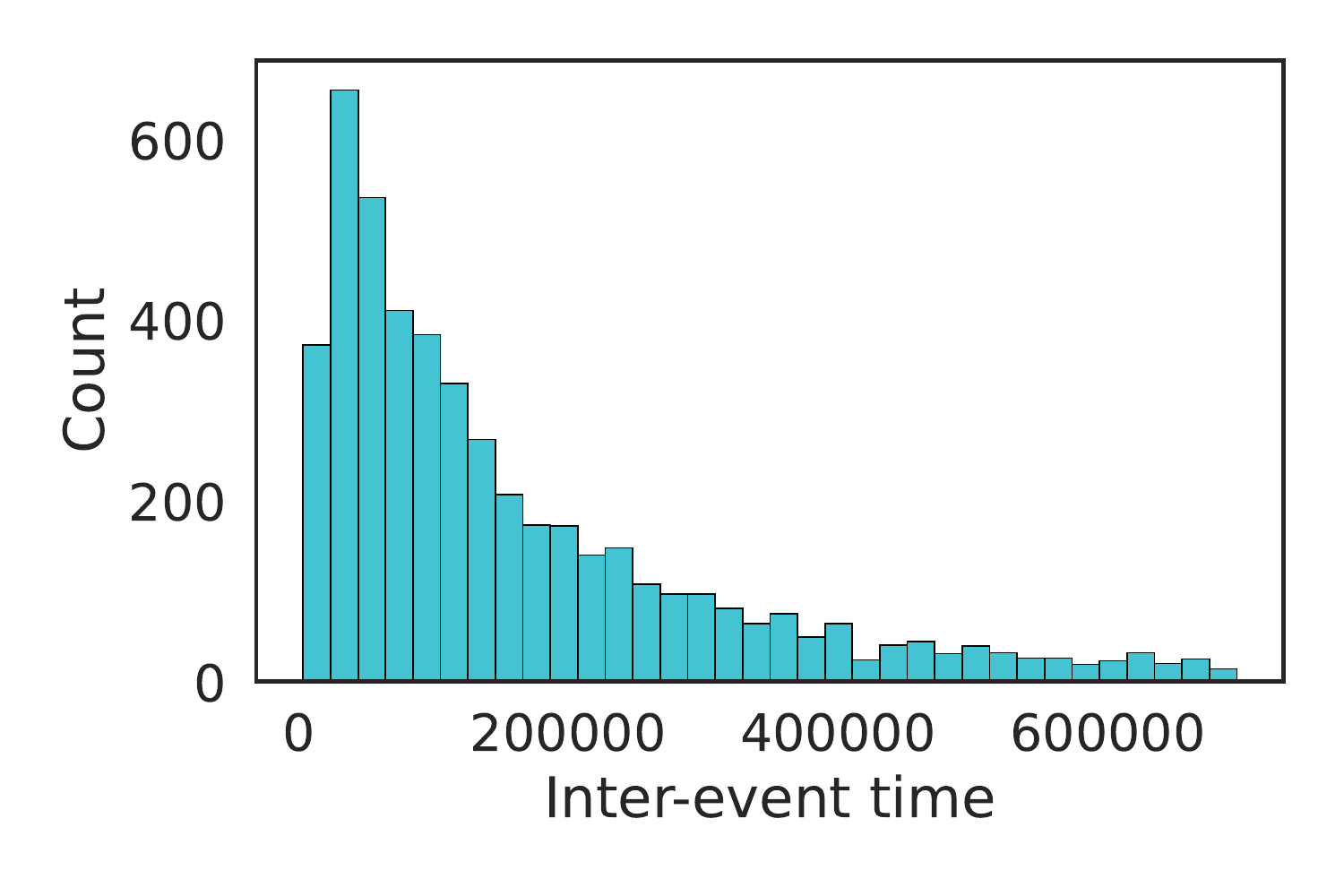}
      
      \caption{Vinted inter-event time distribution.}
    \end{subfigure}

  \caption{Different properties of Stack Overflow and Vinted datasets.}
  \Description{Different properties of Stack Overflow and Vinted datasets.}
      \label{fig:properties}
\end{figure}

The first dataset we use is a simulated Hawkes process. The synthetic Hawkes data serves as a starting point while evaluating selected models' performance. As the data is generated, we can have unlimited samples. Also, as the intensity values at any point in time are known, we are able to compare them with values provided by the trained models. We designed the synthetic dataset consisting of two independent processes. 

The second and the third datasets are real user behavior datasets. Stack Overflow is a question-answering website that awards users various badges based on their activity on the website. The website's users activity is commonly used when benchmarking various temporal point process models \cite{10.1145/2939672.2939875, 1905.10403, 2007.13794, 10.5555/3295222.3295420}. Novel customer behavior dataset comes from a second hand marketplace Vinted, where users sell and buy various goods. 

\begin{figure}[!htbp]
  \centering
    \includegraphics[width=\linewidth]{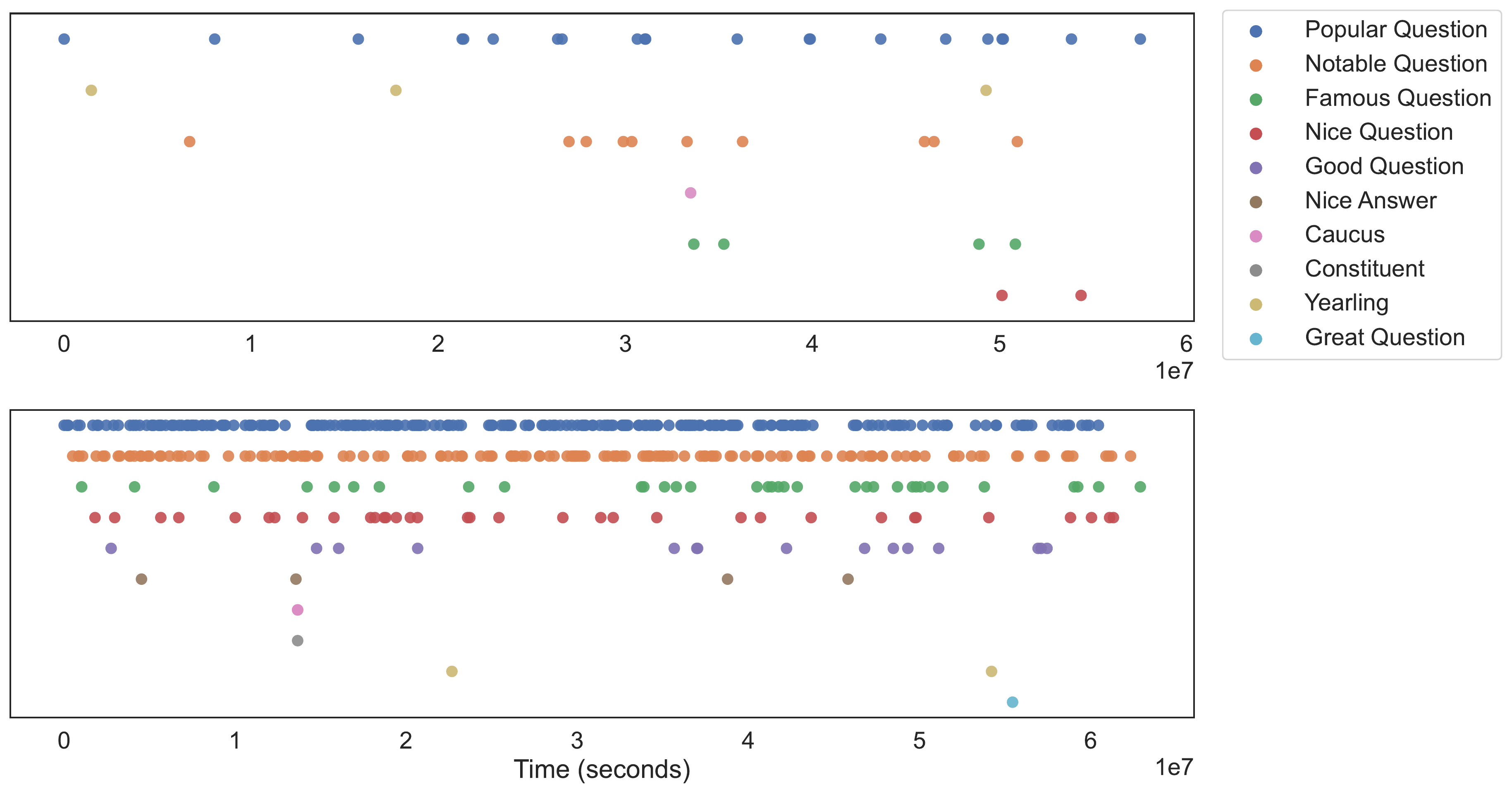}
  \caption{Comparison between passive (top) and active (bottom) Stack Overflow user's activity within the website. Passive website user tends to acquire less diverse badges' set than an active one.}
  \Description{Comparison between active (top) and passive (bottom) Stack Overflow user's activity within the website. Passive website user tends to acquire less diverse badges' set than an active one.}
  \label{fig:so_users_activity}
\end{figure}

\begin{figure}[!htbp]
  \centering
    \includegraphics[width=\linewidth]{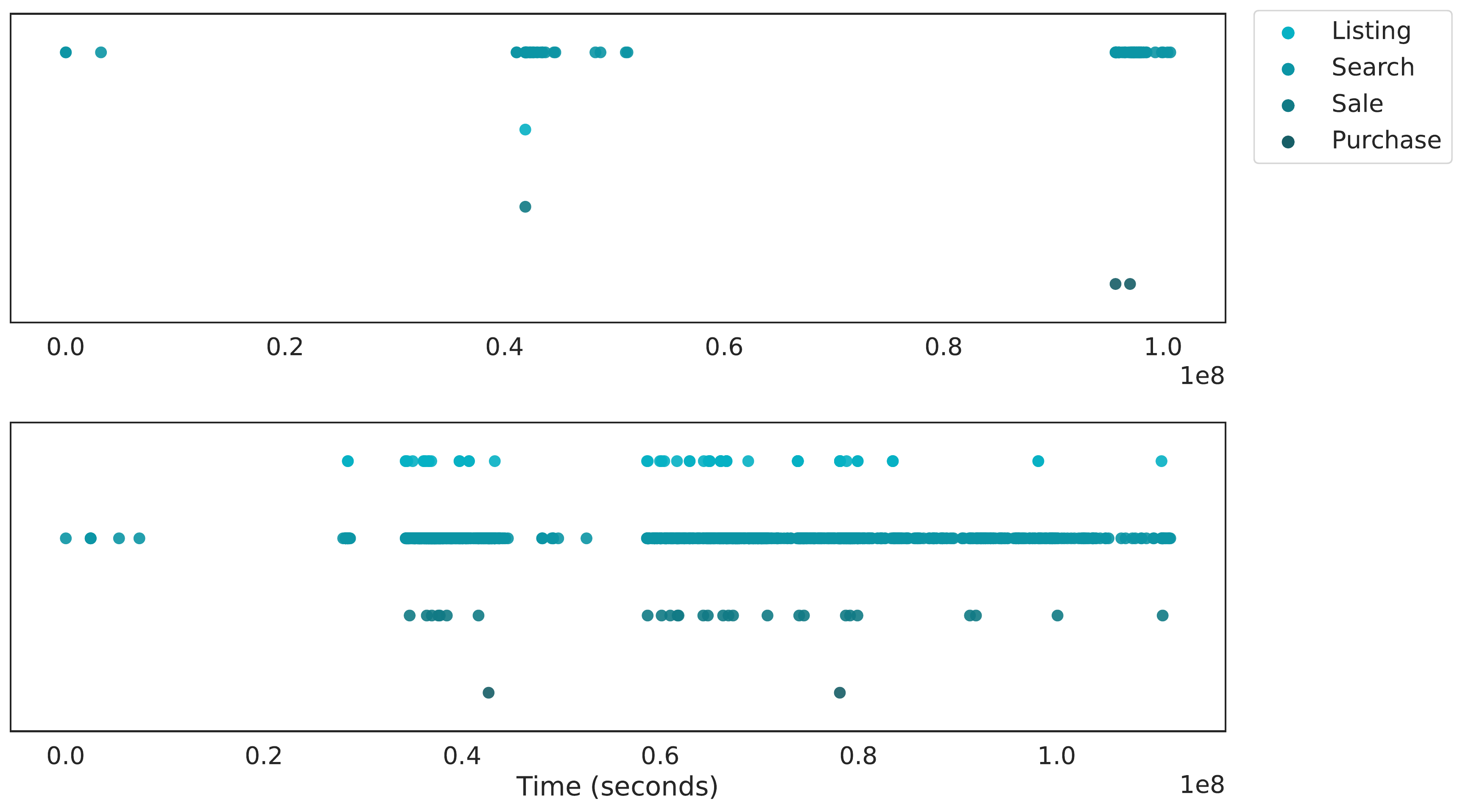}
  \caption{Comparison between passive (top) and active (bottom) Vinted user's activity within the platform. Active Vinted user completes same actions as the passive user, only in a more frequent manner. }
  \Description{Comparison between passive (top) and active (bottom) Vinted user's activity within the platform. Active Vinted user completes same actions as the passive user, only in a more frequent manner. }
  \label{fig:vinted_users_activity}
\end{figure}

We collect Vinted users' activity data by sampling all of the recorded activity (from user's registration to the time of collecting the data) of 5,082 Vinted users. All of the selected users were active in the last year (2021/01/01 - 2022/01/01) and completed at least 40 actions. Inside the dataset, four different types of events can be found: Purchase, Listing, Search, and Sale. Frequent event occurrences and long recorded activity periods result in the dataset being ten times larger than Stack Overflow (4,886,667 vs. 480,414 events). While inspecting the event count distributions (see Figure \ref{fig:properties}), we notice that users' activities within Vinted platform are similar to Stack Overflow, where common events' distributions are shaped similarly. Events collected from Vinted platform as well as from the Stack Overflow website suffer from data imbalance. Events like Publicist, Populist, or Booster are given to the website users only when he accomplishes unique action. These actions can be tricky to capture for temporal point process models, as the data imbalance is quite significant within the dataset. The most common events are Search and Listing. Purchases and Sales are happening much rarer - as they are influenced not only by users' behavior but also by external factors such as the listed item's appeal or Vinted's recommendation system's quality.

Where the differences between Vinted and Stack Overflow platforms begins is the measured inter-event times (see Figure \ref{fig:properties}). Vinted users are returning to the platform more frequently than Stack Overflow ones. This difference is caused by the fact that Stack Overflow browsing activity is not recorded. Activity tendencies between active and passive platform users are also different between mentioned platforms. Active Vinted user completes the same actions as the passive user, only in a more systematic manner (see Figure \ref{fig:vinted_users_activity}). 

Besides user activity records, we collect static Vinted users' features. We use the features to explore a solution to the cold start problem in modeling user behavior. We parameterize temporal point process models in the hope of more accurate event type classification results. 

We split each dataset into training, test and validation parts. Dataset statistics are summarized in Table \ref{tab:dataset_stats}.

\begin{table*}
  \caption{Properties of datasets used for experimentation.}
  \label{tab:dataset_stats}
\begin{tabular}{lllllllll}
\toprule
                 &                     &                    &                    &                      & \multicolumn{4}{c}{Size}                                \\ \cmidrule(l){6-9} 
Dataset & \# classes & Task type & \# events & Avg. length & Train & Valid & Test & Batch \\ \midrule
Hawkes           & 2                   & Multi-class        & 350,281            & 14                   & 233,537        & 58,451         & 58,293        & 512            \\
Stack Overflow   & 22                  & Multi-class        & 480,414            & 72                   & 335,870        & 47,966         & 96,578        & 32             \\
Vinted           & 4                   & Multi-class        & 4,886,657          & 962                  & 3,428,380      & 482,512        & 975,765       & 4              \\ \bottomrule
\end{tabular}
\end{table*}

\subsection{Training}

We base our model selection on the model's performance in capturing the type and time of the next event based on the time it occurred. Also, the accessibility of a source code is a significant deciding factor. We select \cite{1905.10403} and \cite{2007.13794} as the primary source of models used for our experimentation. \cite{2007.13794} demonstrates that models based on Encoder-decoder architectures are effective not only for Natural Language Processing \cite{1406.1078, 1706.03762} but also work well in a temporal point processing field. In the paper, the usage of Self-Attention as a building block of the model's architecture proves to be beneficial while modeling healthcare records. Where \cite{1905.10403} showcases a Neural Jump Stochastic Differential Equations - model that captures temporal point processes with a piecewise-continuous latent trajectory. The model demonstrates its predictive capabilities on various synthetic and real-world marked point process datasets.

We conclude the final list of models from six different ones. Five of them are selected from \cite{2007.13794}: \textbf{SA-COND-POISSON}, \textbf{SA-LNM}, \textbf{SA-MLP-MC}, \textbf{SA-RMTPP-POISSON}, \textbf{SA-SA-MC}. Where the beginning of the model's name "\textit{SA-}" means the type of encoder (Self-Attention), and the rest of the name marks the type of encoder. The sixth model used in experimentation is Neural Jump Stochastic Differential Equations (\textbf{NJSDE}).

For all models, we use the hyperparameters specified in the original literature, with an exception of batch size and training epochs.
Experiments on Vinted data use the same training configuration as experiments on Stack Overflow users' activity.

Furthermore, we explore different solutions to the cold start problem in modeling behavior of new Vinted users, we experiment with parameterizing self-attention based NTPPs. Similarly to \cite{gu-budhkar-2021-package}, we join processed tabular user features to the output of Encoder's transformer network. We train all models on a workstation with a \textit{Intel 32 core 3rd gen Xeon CPU and 120 GB memory}. The complete implementation of our algorithms and experiments are available at \hyperlink{https://github.com/dqmis/ntpps}{https://github.com/dqmis/ntpps}.

\subsection{Model evaluation}
All trained models are trained and evaluated using a five-fold cross-validation strategy. As in \cite{1905.10403}, we use a weighted F1 score as one of the primary evaluation metrics. This allows us to verify model's capabilities of predicting the type of the next event given the time it occurred. As a secondary metric, we use classification accuracy, mainly to compare our results with \cite{2007.13794}. Despite its popularity in quantifying the predictive performance of TPPs, we do not use MAE/MSE metric in this paper. As the metric compares actual intensity values with the model predicted ones, the lack of such data (only synthetic Hawkes dataset intensity values are known) is the main reason that determines our choice. Additionally, we assess each model's predictions via a classification report. As the data imbalance problem is a concern, the metric enables us to validate model's performance in predicting low-samples label.

\section{Results}
\begin{table*}
  \caption{Evaluation on Hawkes and Stack Overflow datasets. Best performances are boldened. Where comparable, results from \cite{1905.10403} and \cite{2007.13794} are displayed in \textit{italic font style}. }
  \label{tab:so_hawkes_results}
  \begin{tabular}{llllll}
\toprule
                  & \multicolumn{2}{c}{Hawkes}    &  & \multicolumn{2}{c}{Stack Overflow} \\ \cmidrule(lr){2-3} \cmidrule(l){5-6} 
Model             & Accuracy    & F1 score      &  & Accuracy       & F1 score        \\ \midrule
NJSDE             & \textbf{.541} & \textbf{.552} &  & \textbf{.548} \textit{(.527)}    & \textbf{.363}   \\
SA-COND-POISSON   & \textbf{.538} & \textbf{.538} &  & .501             & \textbf{.332} \textit{(.326)}  \\
SA-LNM            & .537          & .536          &  & .369             & .319 \textit{(.305)}            \\
SA-MLP-MC         & .536          & .531          &  & .216             & .194 \textit{(.327)}          \\
SA-RMTPP-POISSON  & .526          & .474          &  & .515             & .316 \textit{(.288)}            \\
SA-SA-MC          & .507          & .467          &  & .382             & .305 \textit{(.324)}            \\ \bottomrule
\end{tabular}
\end{table*}

We report our results on Hawkes and Stack Overflow data (table \ref{tab:so_hawkes_results}) and on Vinted data (table \ref{tab:vinted_results}). First, we note that replication of the models' performance results provided by \cite{1905.10403} and \cite{2007.13794} are mostly successful. Notably, the results of SA-MLP-MC model trained on the Stack Overflow dataset are not similar to the ones presented in \cite{1905.10403}. This could be caused by a mismatch in training environments or errors in our or \cite{2007.13794} evaluation process. Notably, the best performing models were NJSDE and SA-COND-POISSON. Both presented great results while classifying synthetic Hawkes events and Stack Overflow users' actions. However, NJSDE model required much more resources and took longer to train to reach high accuracy (15 hours to train NJSDE model vs. 4 hours to train SA-COND-POISSON). NJSDE implementation proposed by \cite{1905.10403} requires time series conversion into grid later used for modeling. This action consumes more memory as the sequence count in the batch grows. Because of this, we could not validate the NJSDE model's performance on Vinted data.

\begin{figure}[!ht]

  \centering
    \begin{subfigure}{\linewidth}
      \includegraphics[width=\linewidth]{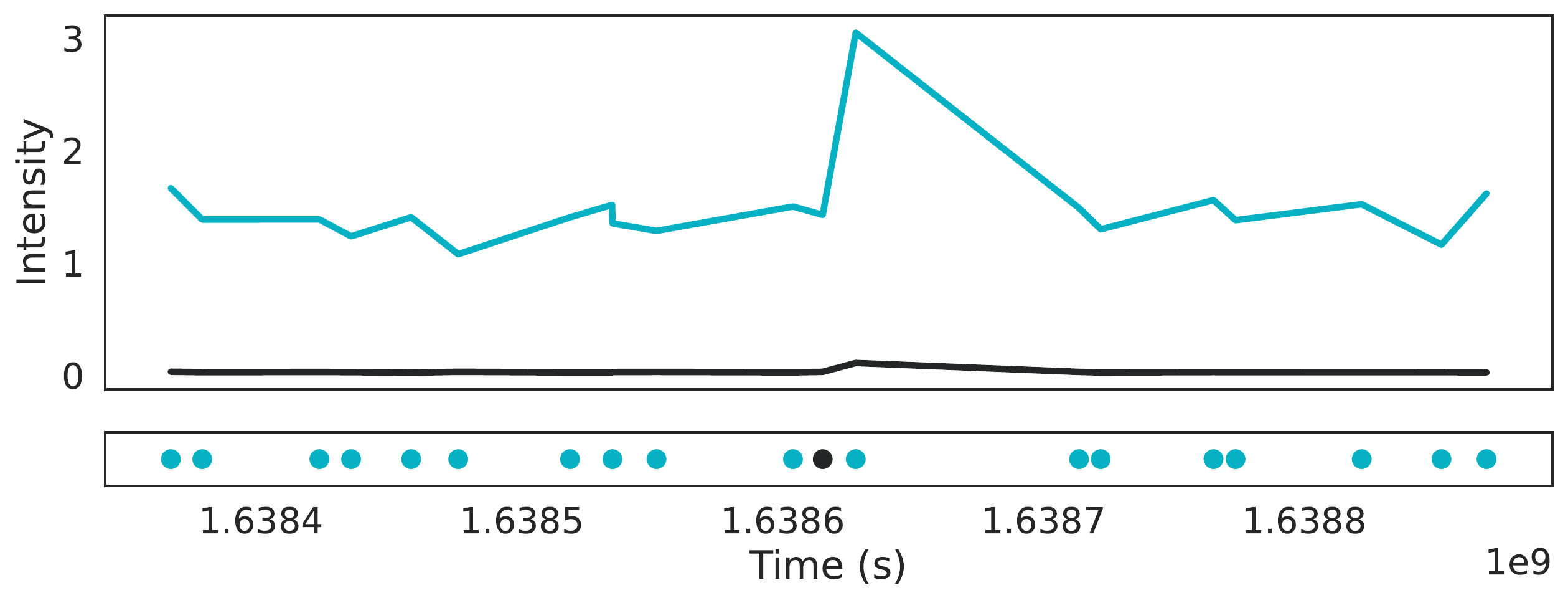}
    \end{subfigure}
  \begin{subfigure}{\linewidth}
      \includegraphics[width=\linewidth]{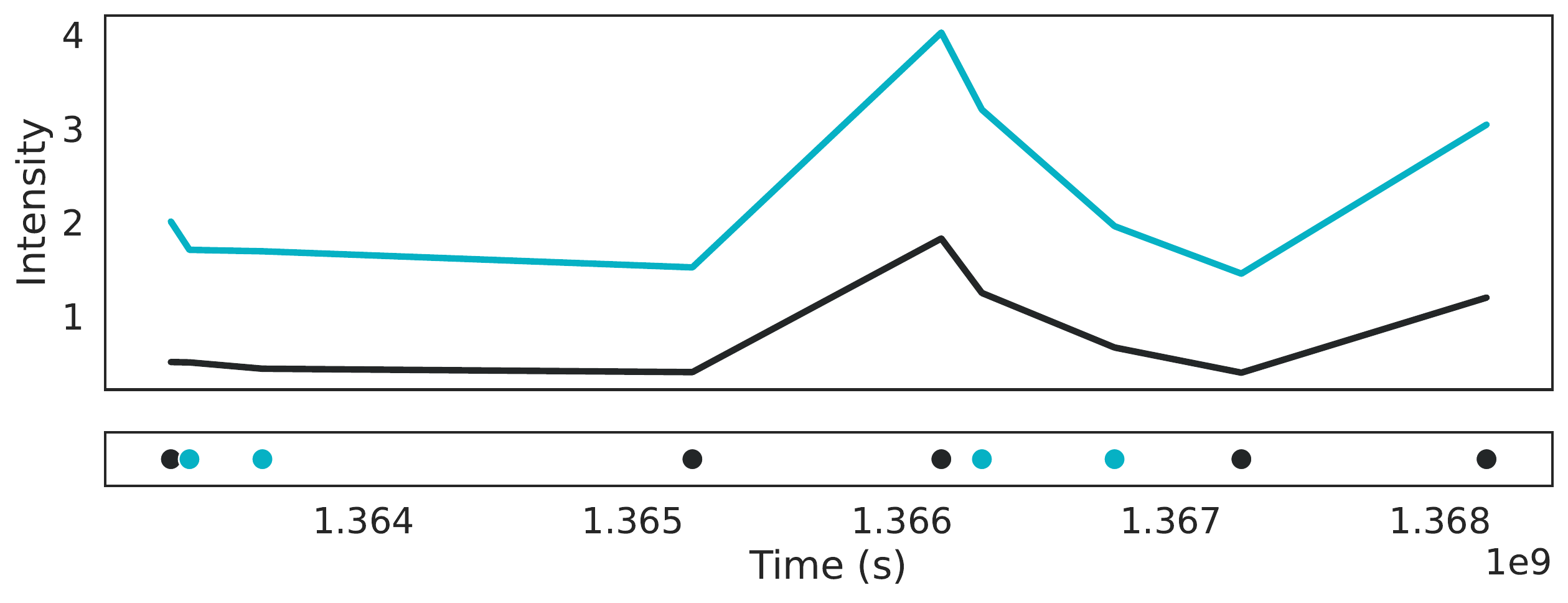}
    \end{subfigure}

  \caption{Intensity functions on several labels provided by \textbf{SA-COND-POISSON} model trained on Vinted (top) and Stack Overflow (bottom) datasets. Common event's intensity (marked green color) is always higher than rare event's one (marked black color). This results in the model omitting rare events while classifying the next event's type.}
  \Description{Intensity functions on several labels provided by \textbf{SA-COND-POISSON} model trained on Vinted (top) and Stack Overflow (bottom) datasets. Dominant class (samples count in the dataset wise) label always has a higher intensity value than the recessive one. This results in the model omitting rare events while classifying the next event's type.}
  \label{fig:intesities}
\end{figure}

Notably, all models, including the best-performing ones, suffer from an inability to classify rare events. This can be seen while comparing the accuracy and F1 score metrics (also see classification report in Table \ref{tab:so_classification}). F1 score and accuracy used in the papers that inspired our research fail to report this phenomenon. This is caused by a significant data imbalance in the Stack Overflow and later discussed Vinted datasets. Also, some types of events could be hard to capture, not only based on their rarity but also their nature. For example, events inside Stack Overflow dataset like Yearling (awarded for being an active user for one year time period) or Caucus (awarded for visiting an election during any phase of an active election and having enough reputation for casting a vote) can be difficult to identify without a time dimension. Also, some of the badges like Promoter (given for the first bounty user offers on his question) are less based on the user's activity and more on external factors like the importance of the asked question.

\begin{table}[!ht]
  \caption{Evaluation on Vinted dataset. The scores are provided in two sections: where model was parameterised with user features / and where it was not. Best performances are boldened.}
  \label{tab:vinted_results}
    \begin{tabular}{llll}
    \toprule
    Model            & Accuracy    & F1 score      \\ \midrule
    SA-LNM           & .842 / .840          & .771 / .768          \\
    SA-COND-POISSON  & \textbf{.882} / \textbf{.882} & \textbf{.857} / \textbf{.856} \\
    SA-SA-MC         & .707 / .839 & .657 / .767        \\
    SA-MLP-MC        & .654 / .643          & .676 / .670          \\
    SA-RMTPP-POISSON & .842 / .839          & .771 / .767          \\ \bottomrule
    \end{tabular}
\end{table}

Same tendencies can be seen while inspecting performance results of models trained on Vinted data. While the best performing SA-COND-POISSON model was able to reach an F1 score of 0.857, it still failed to predict some event types like Purchase and Sale (see Table \ref{tab:vinted_classification}). Interestingly Sale event had twice as many samples as the Listing but failed to be captured by the model. As with Stack Overflow badges, this can be explained by looking at the nature of the event. Searching for a new dress to buy or listing some pair of shoes are events caused mainly by users' habits. Purchases are motivated more by Vinted's search quality or recommendation system. The sale of the uploaded item is also less caused by its seller's past activity and more by the item's quality and appeal to the buyers. Inspecting differences between intensity values of the SA-COND-POISSON model trained on Vinted and Stack Overflow data (see Figure \ref{fig:intesities}) gives a better understanding of the model's behavior. There is a significant difference between the intensity values of a frequent and less frequent event type. The difference tends to stay the same between all time points. This results in the model capturing overall intensity changes but lacking the ability to identify rare events.
Furthermore, our experimentation with parameterizing encoder-decoder architecture-based models with user's static features did not produce valuable results (see Table \ref{tab:vinted_results}). There is one case (SA-SA-MC) where parameterized model performs better than its counterpart, but this tendency is not identified while comparing the rest of the models. The lack of models' performance improvement is probably caused by the fact that selected user features do not directly impact user's behavior. However, to verify if the NTPPs parameterization with tabular features is beneficial, we need to conduct more detailed experiments with a more diverse set of features.

\section{Conclusion}

We discussed broad industry application cases of NTPPs and their ability to capture online platform users' behavior. We used synthetic Hawkes dataset, Stack Overflow users' badges, and second-hand marketplace Vinted user activity as our experimentation datasets. Furthermore, we identify that Self-Attention and Neural Ordinary Differential Equations based NTPP models fail to capture the time of the next rare event but succeed in identifying overall user activity intensity. While the NJSDE model is optimal for identifying the next event's type and time, we note that it is not scalable to extensive industry data. Our experimentation with parametrizing Self-Attention-based NTPPs did not show any significant improvements. However, we identify that we need to conduct more experiments to verify if the method is beneficial. 

\subsection{Broader Impact}
While the explored NTPPs are not suitable for predicting the time of the next rare event, it does not mean they are not beneficial for subscription and non-subscription-based online platforms. These models can be valuable when solving problems such as churn prediction. By grouping individual events into user sessions, we would be able to detect when a general user's intensity is decreasing, thus letting us know when to act. Furthermore, we can optimize recommendation systems or marketing strategies by knowing the time of the next user's arrival on the online platform.

\bibliographystyle{ACM-Reference-Format}
\bibliography{bibliography}

\clearpage

\clearpage

\appendix

\section{Classification reports}

\begin{table}[!htbp]
  \caption{Classification report of \textbf{NJSDE} model predictions on Stack Overflow dataset.}
  \label{tab:so_classification}
\begin{tabular}{llllll}
\toprule
Event type       &  & Precision & Recall & F1 Score & \# samples \\ \midrule
Nice Question    &  & 0         & 0      & 0        & 6,016      \\
Good Answer      &  & 0         & 0      & 0        & 1.939      \\
Guru             &  & 0         & 0      & 0        & 653        \\
Popular Question &  & .471      & .975   & .634     & 42,599     \\
Famous Question  &  & 0         & 0      & 0        & 5,544      \\
Nice Answer      &  & .317      & .511   & .392     & 6,032      \\
Good Question    &  & 0         & 0      & 0        & 1,846      \\
Caucus           &  & 0         & 0      & 0        & 1,788      \\
Notable Question &  & 0         & 0      & 0        & 21,956     \\
Necromancer      &  & 0         & 0      & 0        & 1,308      \\
Promoter         &  & 0         & 0      & 0        & 207        \\
Yearling         &  & 0         & 0      & 0        & 2,400      \\
Revival          &  & 0         & 0      & 0        & 832        \\
Enlightened      &  & 0         & 0      & 0        & 1,859      \\
Great Answer     &  & 0         & 0      & 0        & 251        \\
Populist         &  & 0         & 0      & 0        & 99         \\
Great Question   &  & 0         & 0      & 0        & 269        \\
Constituent      &  & 0         & 0      & 0        & 526        \\
Announcer        &  & 0         & 0      & 0        & 364        \\
Stellar Question &  & 0         & 0      & 0        & 65         \\
Booster          &  & 0         & 0      & 0        & 17         \\
Publicist        &  & 0         & 0      & 0        & 8          \\ \bottomrule
\end{tabular}
\end{table}

\begin{table}[!htbp]
  \caption{Classification report of \textbf{SA-COND-POISSON} model predictions on Vinted dataset.}
  \label{tab:vinted_classification}
  \begin{tabular}{llllll}
\toprule
Event type &  & \multicolumn{1}{l}{Precision} & \multicolumn{1}{l}{Recall} & \multicolumn{1}{l}{F1 score} & \multicolumn{1}{l}{\# samples} \\ \midrule
Listing    &  & .685                          & .654                       & .669                         & 102,297                        \\
Purchase   &  & 0                             & 0                          & 0                            & 23,642                         \\
Sale       &  & 0                             & 0                          & 0                            & 293,68                         \\
Search     &  & .904                          & .968                       & .935                         & 820,458                        \\ \bottomrule
\end{tabular}   
\end{table}

\vfill\eject

\section{Hawkes datasets}
The parameters of our Hawkes datasets are:
$$
  \mu =
  \left[ {\begin{array}{c}
    0.1\\
    0.05 \\
  \end{array} } \right],
  \alpha =
  \left[ {\begin{array}{cc}
    0.2 & 0.0 \\
    0.0 & 0.4 \\
  \end{array} } \right],
    \beta =
  \left[ {\begin{array}{cc}
    1.0 & 1.0 \\
    1.0 & 1.0 \\
  \end{array} } \right]
$$

\end{document}